\documentclass[runningheads]{llncs}
\usepackage{graphicx}
\graphicspath{ {images/} }
\usepackage{amsmath,amssymb} 
\usepackage{color}
\usepackage{tabularx}
\usepackage{placeins}
\usepackage[strings]{underscore}

\usepackage{hyperref}

\DeclareMathOperator{\arctantwo}{arctan2}
\newcolumntype{Z}{ >{\centering\arraybackslash}X}

\begin{document}

\title{3DFeat-Net: Weakly Supervised Local 3D Features for Point Cloud Registration} 

\titlerunning{3DFeat-Net}

\author{Zi Jian Yew \and Gim Hee Lee}
%
\authorrunning{Zi Jian Yew and Gim Hee Lee}

\institute{Department of Computer Science, National University of Singapore \\
\email{\{zijian.yew, gimhee.lee\}@comp.nus.edu.sg}}
\maketitle

\begin{abstract}
In this paper, we propose the 3DFeat-Net which learns both 3D feature detector and descriptor for point cloud matching using weak supervision. Unlike many existing works, we do not require manual annotation of matching point clusters. Instead, we leverage on alignment and attention mechanisms to learn feature correspondences from GPS/INS tagged 3D point clouds without explicitly specifying them. We create training and benchmark outdoor Lidar datasets, and experiments show that 3DFeat-Net obtains state-of-the-art performance on these gravity-aligned datasets.
\end{abstract}

\section{Introduction}

\begin{figure}[t]
\centering
\includegraphics[height=1.5in]{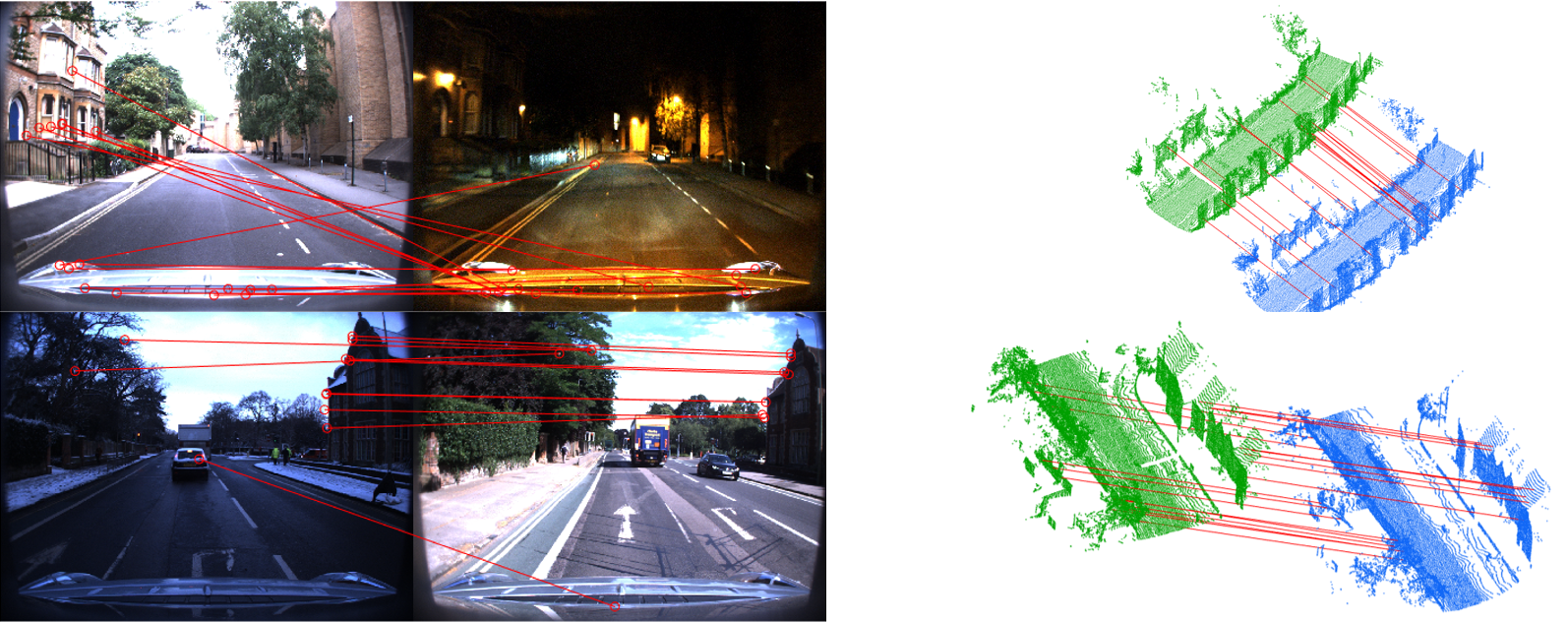}
\caption{Left: 2D images from the Oxford dataset at different times of the day (top) or seasons of the year (bottom) give mostly wrong matches even after RANSAC. Right: 3D point cloud of the same scene remains largely the same and can be matched easily.}
\label{fig:visual-vs-lidar}
\end{figure}

3D point cloud registration plays an important role in many real-world applications such as 3D Lidar-based mapping and localization for autonomous robots, and 3D model acquisition for archaeological studies, geo-surveying and architectural inspections etc. Compared to images, point clouds exhibit less variation and can be matched under strong lighting changes, i.e. day and night, or summer and winter (Fig. \ref{fig:visual-vs-lidar}).
A two-step process is commonly used to solve the point cloud registration problem - (1) establishing 3D-3D point correspondences between the source and target point clouds, and (2) finding the optimal rigid transformation between the two point clouds that minimizes the total Euclidean distance between all point correspondences. Unfortunately, the critical step of establishing 3D-3D point correspondences is non-trivial. Even though many handcrafted 3D feature detectors \cite{ISS,LSP} and descriptors \cite{PFH,FPFH,SpinImage,USC,SHOT} have been proposed over the years, the performance of establishing 3D-3D point correspondences remains unsatisfactory. As a result, iterative algorithms, e.g. Iterative Closest Point (ICP) \cite{ICP}, that circumvent the need for wide-baseline 3D-3D point correspondences with good initialization and nearest neighbors, are often used. This severely limits usage in applications such as global localization / pose estimation \cite{LeePose15} and loop-closures \cite{dube2016segmatch} that require wide-baseline correspondences.

Inspired by the success of deep learning for computer vision tasks such as image-based object recognition \cite{SIFTObjectRecognition}, several deep learning based works that learn 3D feature descriptors for finding wide-baseline 3D-3D point matches have been proposed in the recent years. Regardless of the improvements of these deep learning based 3D descriptors over the traditional handcrafted 3D descriptors, none of them proposed a full pipeline that uses deep learning to concurrently learn both the 3D feature detector and descriptor. This is because the existing deep learning approaches are mostly based on supervised learning that requires huge amounts of hand-labeled data for training. It is impossible for anyone to manually identify and label salient 3D features from a point cloud. Hence, most existing approaches focused only on learning the 3D descriptors, while the detection of the 3D features are done with random selection \cite{zeng20163dmatch,Elbaz2017CVPR}. On the other hand, it is interesting to note the availability of an abundance of GPS/INS tagged 3D point cloud based datasets collected over large environments, e.g. the Oxford RobotCar \cite{RobotCarDatasetIJRR} and KITTI \cite{kitti} datasets etc. This naturally leads us into the question: ``Can we design a deep learning framework that concurrently learns the 3D feature detector and descriptor from the GPS/INS tagged 3D point clouds?"

In view of the difficulty to get datasets of accurately labeled salient 3D features for training the deep networks, we propose a \textit{weakly supervised} deep learning framework - the 3DFeat-Net to holistically learn a 3D feature detector and descriptor from GPS/INS tagged 3D point clouds. Specifically, our 3DFeat-Net is a Siamese architecture \cite{bromley1994siamese} that learns to recognize whether two given 3D point clouds are taken from the same location. 
We leverage on the recently proposed PointNet \cite{pointnet,pointnetpp} to enable us to directly use the 3D point cloud as input to our network. The output of our 3DFeat-Net is a set of local descriptor vectors. The network is trained by minimizing a Triplet loss \cite{schroff2015facenet} where the positive and ``hardest" negative samples are chosen from the similarity measures between all pairs of descriptors \cite{karpathy2015alignment} from two input point clouds. 
Furthermore, we add an attention layer \cite{delf} that learns importance weights that weigh the contribution of each input descriptor towards the Triplet loss. During inference, we use the output from the attention layer to determine the saliency likelihood of an input 3D point. Additionally, we take the output descriptor vector from our network as the descriptor for finding good 3D-3D correspondences. Experimental results from real-world datasets \cite{RobotCarDatasetIJRR,kitti,ethdata} validates the feasibility of our 3DFeat-Net. 

\medskip
\noindent \textbf{Our contributions} in this paper can be summarized as follows:
\begin{itemize}
  \item Propose a \textit{weakly supervised} network that holistically learns a 3D feature detector and descriptor using only GPS/INS tagged 3D point clouds.
  \item Use an attention layer \cite{delf} that allows our network to learn the saliency likelihood of an input 3D point. 
  \item Create training and benchmark datasets from Oxford RobotCar dataset \cite{RobotCarDatasetIJRR}. 
\end{itemize}

\noindent We have made our source code and dataset available online.\footnote{\url{https://github.com/yewzijian/3DFeatNet}}

\section{Related Work}
Existing approaches of the local 3D feature detectors and descriptors are heavily influenced by the widely studied 2D local features methods \cite{SIFT,SURF}, and can be broadly categorized into handcrafted \cite{ISS,LSP,PFH,FPFH,SpinImage,USC,SHOT} and learning approaches \cite{zeng20163dmatch,Elbaz2017CVPR,CGF,DescriptorSpecificDetector15,ppfnet} - i.e. pre- and post- deep learning approaches.

\medskip
\noindent \textbf{Handcrafted 3D Features} \enskip
Several handcrafted 3D features are proposed before the popularity of deep learning. The design of these features are largely based on the domain knowledge of 3D point clouds by the researchers.
The authors of \cite{ISS} and \cite{LSP} detects salient keypoints which have large variations in the principal direction \cite{ISS}, or unique curvatures \cite{LSP}. 
The similarity between keypoints can then be estimated using descriptors. PFH \cite{PFH} and FPFH \cite{FPFH} consider pairwise combinations of surface normals to describe the curvature around a keypoint. Other 3D descriptors \cite{SpinImage,USC} build histograms based on the number of points falling into each spatial bin around the keypoint. A comprehensive evaluation of the common handcrafted 3D detectors and descriptors can be found in \cite{HandcraftedComparison}. As we show in our evaluation, many of these handcrafted detectors and descriptors do not work well on real world point clouds, which can be noisy and low density.

\medskip
\noindent \textbf{Learned 2D Features} \enskip
Some recent works have applied deep learning to learn detectors and descriptors on images for 2D-2D correspondences. LIFT \cite{yi2016lift} learns to distinguish between matching and non-matching pairs with a Siamese CNN, 
where the matching pairs are obtained from feature matches that survive the Structure from Motion (SfM) pipeline. TILDE \cite{tilde2015} learns to detect keypoints that can be reliably matched over different lighting conditions. These works rely on the matches provided by handcrafted 2D features, e.g. SIFT \cite{SIFT}, but unfortunately handcrafted 3D features are less robust and do not provide as good a starting point to learn better features. On the other hand, a recently proposed work - DELF \cite{delf} uses a weakly supervised framework to learn salient local 2D image features through an attention mechanism in a landmark classification task. This motivates our work in using an attention mechanism to identify good 3D local keypoints and descriptors for matching. 

\medskip
\noindent \textbf{Learned 3D Features} \enskip
The increasing success and popularity of deep learning has recently inspired many learned 3D features.
3DMatch \cite{zeng20163dmatch} uses a 3D convolutional network to learn local descriptors for indoor RGB-D matching by training on matching and non-matching pairs. PPFNet \cite{ppfnet} operates on raw points, incorporating point pair features and global context to improve the descriptor representation. Other works such as CGF \cite{CGF} and LORAX \cite{Elbaz2017CVPR} utilize deep learning to reduce the dimension of their handcrafted descriptors.
Despite the good performance of these works, none of them learns to detect keypoints. The descriptors are either computed on all or random sampled points. On the other hand, \cite{DescriptorSpecificDetector15} learns to detect keypoints that give good matching performance with handcrafted SHOT \cite{SHOT} descriptors. This provides the intuition for our work, i.e. a good keypoint is one that gives a good descriptor performance.

\section{Problem Formulation}
A point cloud $P$ is represented as a set of $N$ 3D points $\{x_i | i = 1, ..., N\}$. 
Each point cloud $P^{(m)}$ is cropped to a ball with fixed radius $R$ around its centroid $c_m$. We assume the absolute pose of the point cloud is available during training, e.g. from GPS/INS, but is not sufficiently accurate to infer point-to-point correspondences. We define the distance between two point clouds $d(m,n)$ as the Euclidean distance between their centroids, i.e. $d(m,n) = \|c_m - c_n\|_2$. 

We train our network using a set of triplets containing an anchor, positive and negative point cloud $\{P^{(anc)}, P^{(pos)},$ $P^{(neg)}\}$, similar to typical instance retrieval approaches \cite{arandjelovic2016netvlad,gordo2016deepimageretrieval}. 
We define positive instances as point clouds with distance to the anchor below a threshold, $d(anc, pos) < \tau_p$. Similarly, negative instances are point clouds far away from the anchor, i.e. $d(anc, neg) > \tau_n$. The thresholds $\tau_p$ and $\tau_n$ are chosen such that positive and negative point clouds have large and small overlaps with the anchor point cloud respectively.

The objective of our network is to learn to find a set of correspondences $\{(x_{1}^{(m)}, x_{1}^{(n)}), (x_{2}^{(m)}, x_{2}^{(n)}), ..., (x_{L}^{(m)}, x_{L}^{(n)}) | x_{i}^{(m)} \in P^{(m)}, x_{j}^{(n)} \in P^{(n)}\}$ between a subset of points in two point clouds $P^{(m)}$ and $P^{(n)}$. Our network learning is weakly supervised in two ways. Firstly, only model level annotations in the form of relative poses of the point clouds are provided, and we do not explicitly specify which subset of points to choose for the 3D features. Secondly, the ground truth poses are not accurate enough to infer point-to-point correspondences.

\section{Our 3DFeat-Net}

\subsection{Network Architecture}
\begin{figure}[t]
\centering
\includegraphics[width=\textwidth]{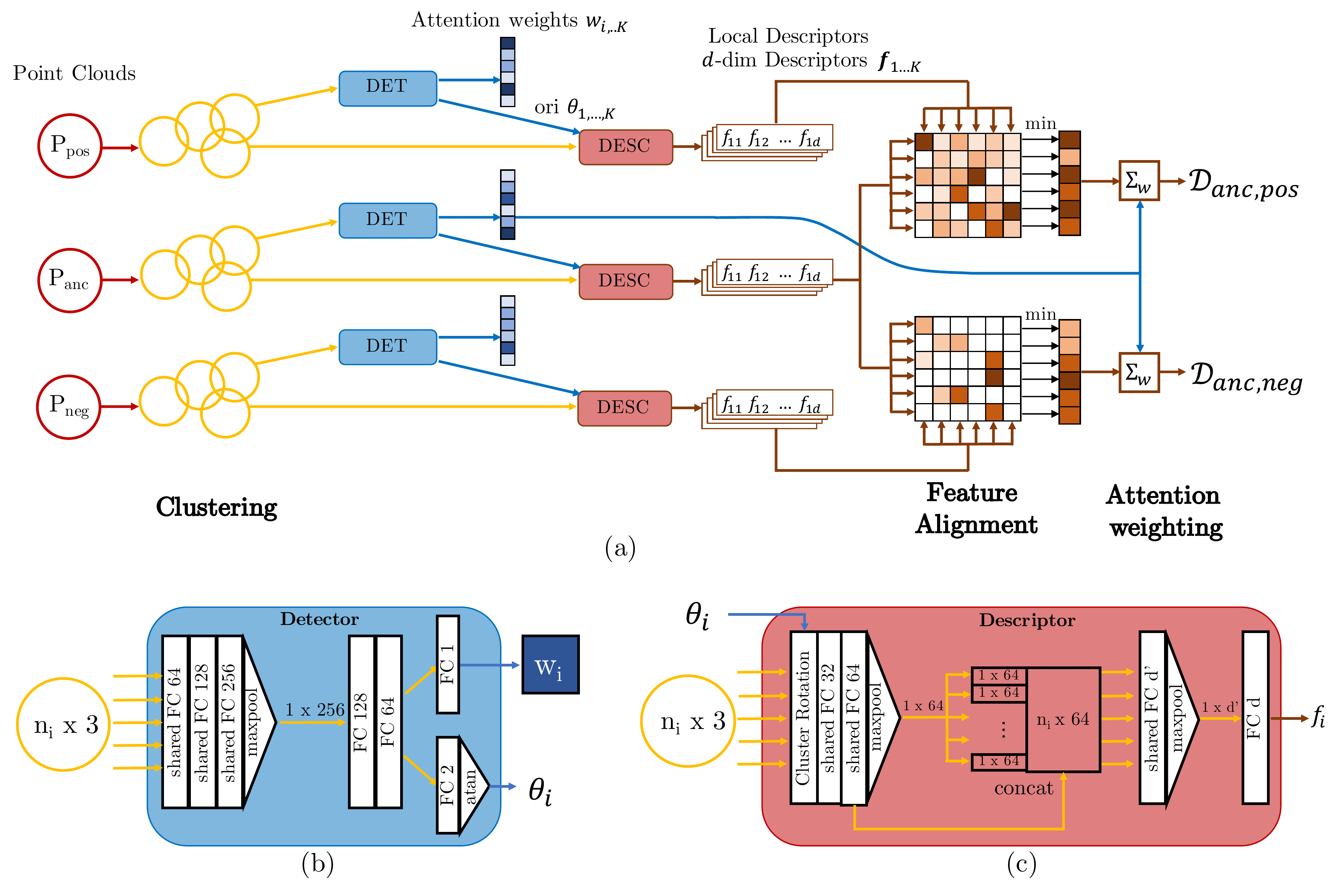}
\caption{(a) Network architecture of our 3DFeat-Net. The three-branch Siamese architecture uses a training tuple of an anchor, positive and negative point cloud to compute a triplet loss. (b) and (c) show detailed view of the detector and descriptor networks.}
\label{fig:architecture}
\end{figure}

Fig. \ref{fig:architecture} shows the three-branch Siamese architecture of our 3DFeat-Net. Each branch takes an entire point cloud $P$ as input. Point clusters $\{C_1, ..., C_K\}$ are sampled from the point cloud in a clustering layer.
For each cluster $C_k$, an orientation $\theta_k$ and attention $w_k$ \cite{delf} are predicted by a detector network.
A descriptor network then rotates the cluster $C_k$ to a canonical configuration
using the predicted orientation $\theta_k$ and computes a descriptor $f_k \in \mathbb{R}^d$.

We train our network with the triplet loss to minimize the difference between the anchor and positive point clouds, while maximizing the difference between anchor and negative point clouds. To allow the loss to take individual cluster descriptors  into account, we use an alignment model \cite{karpathy2015alignment} to align each descriptor to its best match before aggregating the loss. Since not all sampled clusters have the same distinctiveness, the predicted attention $w_k$ from the detector is used to weigh the contribution of each cluster descriptor in the training loss. These attention weights are learned on arbitrarily sampled clusters during training, and later used to detect distinctive keypoints in the point cloud during inference.

\subsubsection{Clustering}
The first stage of the network is to sample clusters from the point cloud. To this end, we use the sample and grouping layers in PointNet++ \cite{pointnetpp}. The sampling layer samples a set of points $\{x_{i_1}, x_{i_2}, \ldots, x_{i_K}\}$ 
from an input point cloud $P$. The coordinates of these sampled points and the point cloud are then passed into the grouping layer that outputs $K$ clusters of points.
Each cluster $C_k$ is a collection of points in a local region of a predefined radius $r_{cluster}$ around the sampled point $x_{i_k}$. These clusters are used as support regions to compute local descriptors, analogous to 2D image patches around detected keypoints in 2D feature matching frameworks. We use the iterative farthest point sampling scheme as in PointNet++ for sampling, but any form of sampling that can sufficiently cover the point cloud (e.g. Random Sphere Cover Set \cite{Elbaz2017CVPR}) is also suitable. Such sampling schemes increases the likelihood that each sampled cluster in an anchor point cloud has a nearby cluster in the positive point cloud.

\subsubsection{Detector}
Each cluster $C_k$ sampled by the clustering step is passed to the detector network that predicts an orientation $\theta_k$ and an attention score $w_k$. The attention score $w_k$ is a positive number that indicates the saliency of the input cluster $C_k$. This design is inspired by typical 2D feature detectors, e.g. SIFT \cite{SIFT}. The predicted orientation is used to rotate the cluster to a canonical orientation, so that the final descriptor is invariant to the cluster's original orientation.

We construct our detector network using point symmetric functions defined in PointNet \cite{pointnet}, which is defined as $f(\{x_1, ..., x_n\}) = g(h(x_1), ..., h(x_n))$, where $h(.)$ is a shared function that transforms each individual point $x_i$, and $g(.)$ is a symmetric function on all transformed elements. These functions are invariant to point ordering and generates fixed length features given arbitrary sized point clouds. We use a three fully connected layers (64-128-256 nodes) for the implementation of $h(.)$. The symmetric function $g(.)$ is implemented as a max-pooling followed by two fully connected layers (128-64 nodes), before branching into two 1-layer paths for orientation and attention predictions.

We only predict a single 1D rotation angle $\theta_i$, avoiding unnecessary equivariances to retain higher discriminating power. This is reasonable since point clouds are usually aligned to the gravity direction due to the sensor setup (e.g. a Velodyne Lidar mounted upright on a car); for other cases, the gravity vector obtained from an IMU can be used to rotate the point clouds into the upright orientation. Similar to \cite{yi2016orientation}, we do not regress the angle directly. Instead, we regress two separate values $\theta_{k_1}$ and $\theta_{k_2}$ that denote the sine and cosine of the angle. We use a $\ell_2$ normalization layer to add a constraint of $\theta_{k_1}^2 + \theta_{k_2}^2 = 1$ to ensure valid sine and cosine values. The final angle can be computed as $\theta_k = \arctantwo(\theta_{k_1}, \theta_{k_2})$. 
For the attention weights $w_i$'s, we use the softplus activation as suggested by \cite{delf} to prevent the network from learning negative attention weights.

\subsubsection{Descriptor}
Our descriptor network takes each cluster $C_k$ from the clustering layer and orientation $\theta_k$ from the detector network as inputs, and generates a descriptor $f_k \in \mathbb{R}^d$ for each cluster.
More specifically, $\theta_k$ is first used to rotate cluster $C_k$ into a canonical configuration, before they are passed into another point symmetric function to generate the descriptor.
In practice, we find it helpful to aggregate contextual information in the computation of the descriptor. Hence, after applying max-pooling to obtain a cluster feature vector, we concatenate this cluster feature vector with the individual point features to incorporate context. We then apply a single fully connected layer with $d'$ nodes before another max-pooling. Finally, we apply another fully connected layer and $l_2$ normalization to produce a final cluster descriptor $f_k  \in \mathbb{R}^d$ for cluster $C_k$. The addition of the contextual information improves the discriminating power of the descriptor.

\subsubsection{Feature Alignment Triplet Loss}
The output from the descriptor network in the previous stage is a set of features $f_i$ for each cluster. We use the alignment objective introduced in \cite{karpathy2015alignment} to compare individual descriptors since the supervision is given as model-level annotations. Instead of the dot product similarity measure used in \cite{karpathy2015alignment}, we adopt the Euclidean distance measure which is more commonly used for comparing feature descriptors. Specifically, the distance between all pairs of descriptors between the two point clouds
$P^{(m)}$ and $P^{(n)}$ with clusters $\mathbf{C}^{(m)}$ and $\mathbf{C}^{(n)}$ is given by: 
\begin{align}
    \mathcal{D}_{m,n} = \sum_{C_i \in \mathbf{C}^{(m)}} \left( w_i' \cdot \min_{C_j \in \mathbf{C}^{(n)}} \| f_{i} - f_{j} \|_2\right),~~~ 
     w_i' = \frac{w_{i}}{\sum_{j \in P^{(m)}} w_j},
\end{align}

\noindent where $w_i'$ is the normalized attention weight.
Under this formulation, every descriptor from the first point cloud aligns to its closest descriptor in the second one. Intuitively, in a matching point cloud pair, clusters in the first point cloud should have a similar cluster in the second point cloud. For non-matching pairs, the above distance simply aligns a descriptor to one which is most similar to itself, i.e. its hardest negative. This consideration of the hardest negative descriptor in the non-matching image provides the advantage that no explicit mining for hard negatives is required. Our model trains well with randomly sampled negative point clouds.
We formulate the triplet loss for each training tuple $\{P^{(anc)}, P^{(pos)}, P^{(neg)}\}$
as:
\begin{align}
\mathcal{L}_{triplet} = [\mathcal{D}_{anc, pos} - \mathcal{D}_{anc, neg} + \gamma]_{+}, 
\end{align}
\noindent where $[z]_{+} = max(z, 0)$ denotes the hinge loss, and $\gamma$ denotes the margin which is enforced between the positive and negative pairs.

\subsection{Inference Pipeline}
The keypoints and descriptors are computed in two separate stages during inference. In the first stage, the attention scores of all points in a given point cloud are computed. We apply non-maximal suppression over a fixed radius $r_{nms}$ around each point, and keep the remaining $M$ points with the highest attention weights. Furthermore, points with low attention are filtered out by rejecting those with attention $w_i < \beta * \max(w_1, w_2, ..., w_N)$. The remaining points are our detected keypoints. In the next stage, the descriptor network computes the descriptors only for these keypoints. 
The separation of the inference into two detector-descriptor stages is computationally and memory efficient since only the detector network needs to process all the points while the descriptor network processes only the clusters that correspond to the selected keypoints. As a result, our network can scale up to larger point clouds.

\section{Evaluations and Results}
\subsection{Benchmark Datasets}

\subsubsection{Oxford RobotCar}
We use the open-source Oxford RobotCar dataset \cite{RobotCarDatasetIJRR} for training and evaluation. This dataset consists of a large number of traversals over the same route in central Oxford, UK, at different times over a year. 
The push-broom 2D scans are accumulated into 3D point clouds using the associated GPS/INS poses.
For each traversal, we create a 3D point cloud with 30m radius for every 10m interval whenever good GPS/INS poses are available.
Each point cloud is then downsampled using a VoxelGrid filter with a grid size of 0.2m. We split the data into two disjoint sets for training and testing. The first 35 traversals are used for training and the last 5 traversals are used for testing. 
We obtain a total of 21,875 training and initially 828 testing point cloud sets.

We make use of the pairwise relative poses computed from the GPS/INS poses as ground truth to evaluate the performance on the test set.
However, the GPS/INS poses may contain errors in the order of several meters. To improve the fidelity of our test set, we refine these poses using ICP \cite{ICP}. We set one of the test traversals as reference, and register all test point clouds within 10m to their respective point clouds in the reference. We retain matches with an estimated Root Mean Square Error (RMSE) of $<0.5$m, and perform manual visual inspection to filter out bad matches. We get 794/828 good point clouds that give us 3426 pairwise relative poses for testing.
Lastly, we randomly rotate each point cloud around the vertical axis to evaluate rotational equivariances and randomly downsample each test point cloud to 16,384 points.

\begin{table}
\centering
\caption{Breakdown of Oxford RobotCar Data for Training and Testing}
\label{table:data-breakdown}
\begin{tabular*}{0.9\textwidth}{@{\extracolsep{\fill}}c||ccc}
\hline
& \# Traversals & \# Point clouds & \# Matched pairs \\
\hline
Train & 35 & 21875 & -\\
\hline
Test & 5 & 828 & -\\
Test (after registration) & 5 & 794 & 3426\\
\hline
\end{tabular*}
\end{table}

\subsubsection{KITTI Dataset}
We evaluate the performance of our trained network on the 11 training sequences of the KITTI Odometry dataset \cite{kitti} to understand how well our trained detector and descriptor generalizes to point clouds captured in a different city using a different sensor. The KITTI dataset contains point clouds captured on a Velodyne-64 3D Lidar scaner in Karlsruhe, Germany. We sample the Lidar scans at 10m intervals to obtain 2369 point clouds, and downsample them using a VoxelGrid filter with a grid size of 0.2m. We consider the 2831 point cloud pairs that are captured within 10m range of each other. We use the provided GPS/INS pose as ground truth.

\subsubsection{ETH Dataset}
Our network is also evaluated on the ``challenging dataset for point cloud registration algorithms"  \cite{ethdata}. This dataset is captured on a ground Lidar scanner, and contains largely unstructured vegetation unlike the previous two datasets. Following \cite{Elbaz2017CVPR}, we accumulate point clouds captured in one season of the Gazebo and Wood scenes to build a global point cloud, and register local point clouds captured in the other season to it. We take the liberty to build the global point cloud for both scenes using the summer data since \cite{Elbaz2017CVPR} did not state the season that was used.
During pre-procesing, we downsample the point clouds using a VoxelGrid filter of 0.1m grid size. We choose a finer resolution because of the finer features in the vegetation in this dataset.

\subsection{Experimental Setup}
We train using a batch size of 6 triplets, and use the ADAM optimizer with a constant learning rate of 1e-5. We use points within a radius of $R=20$m from the centroid of the point cloud for training, and sample $K=512$ clusters with a radius of $r_{cluster}=2.0$m. The thresholds for defining positive and negative point clouds are set to $\tau_{p}=5$m and $\tau_{n}=50$m. We randomly downsample each point cloud to 4096 points on the fly during training. 
We found that training with this random input dropout leads to better generalization behavior as also observed by \cite{pointnetpp}. We apply the following data augmentations during training time: random jitter to the individual points, random shifts and small rotations. Random rotations are also applied around the z-axis, i.e. upright axis in order to learn rotation equivariance. Note that our training data is already oriented in the upright direction using its GPS/INS pose.

Our network is end-to-end differentiable, but in practice, we find it sometimes hard to train. Hence, we train in two phases to improve stability. We first pretrain the network without the detector for 2 epochs, i.e. the clusters are fed directly into the descriptor network without rotation, and all clusters have equal attention weights. During this phase, we apply all data augmentations except for large 1D rotations. We use these learned weights to initialize the descriptor in the second phase, where we train the entire network and apply all the above data augmentation. Training took 34 hours on a Nvidia Geforce GTX1080Ti.

For inference, we use the following parameters for both the Oxford and KITTI odometry datasets: $r_{nms}=0.5$m, $ \beta=0.01$. For the ETH dataset, we use $r_{nms}=0.3$m, $ \beta=0.005$ to boost the number of detected keypoints in the semi-structured environments. We limit the number of keypoints $M$ to 1024 for all cases, except for the global model in the ETH dataset which we use 2048 due to its larger size.  Inference took around 0.8s for a point cloud with 16,384 points.

\subsection{Baseline Algorithms}
We compare our algorithm with three commonly used handcrafted 3D feature descriptors: Fast Point Feature Histograms (FPFH) \cite{FPFH} (33 dimensions), SpinImage (SI) \cite{SpinImage} (153 dimensions), and Unique Shape Context (USC) \cite{USC} (1980 dimensions). We use the implementation provided in the Point Cloud Library (PCL) \cite{PCL} for all the handcrafted descriptors. 
In addition, we include two recent learned 3D descriptors: Compact Geometric Features (CGF) \cite{CGF} (32 dimensions) and 3DMatch \cite{zeng20163dmatch} (512 dimensions) in our comparisons. Note that our comparisons are done using their provided weights that were pretrained on indoor datasets. This is because we are unable to train CGF and 3DMatch on the weakly supervised Oxford dataset as these networks require strong supervision.
We also train a modified PointNet++ (PN++) \cite{pointnetpp} in a weakly supervised manner on a retrieval task, which we use as a baseline to show the importance of descriptor alignment in learning local descriptors. We modify PointNet++ as follows: the first set abstraction layer is replaced by our detection and description network. The second and third set abstraction layers remain unchanged. We replace the subsequent fully connected layers with fc1024-dropout-fc512 layers. During inference, we extract descriptors as the output from the first set abstraction layer. We tuned the parameters for all descriptors to the best of our ability, except in Section \ref{Descriptor-matching}, where we used the same cluster radii for all descriptors to ensure all descriptors ``see" the same information.

Since none of the above baseline feature descriptors comes with a feature detector, we use the handcrafted detector from Intrinsic Shape Signatures (ISS) \cite{ISS} implemented in PCL. Following \cite{zeng20163dmatch}, we also show the performance on randomly sampled keypoints for 3DMatch.

\subsection{Descriptor Matching}\label{Descriptor-matching}
We first evaluate the ability of the descriptors to distinguish between different clusters using the procedure in \cite{zeng20163dmatch}. We extract each matching cluster at randomly selected locations from two matching model pairs. Non-matching clusters are extracted from two random point clouds at locations that are at least 20m apart. We extract 30,000 pairs of 3D clusters, equally split between matches and non-matches. As in \cite{zeng20163dmatch}, our evaluation metric is the false-positive rate at 95\% recall.
To ensure all descriptors have access to similar amounts of information, we fixed the cluster radius $r_{cluster}$ to 2.0m for all descriptors in this experiment.

\begin{figure}
\centering
\includegraphics[width=0.7\textwidth]{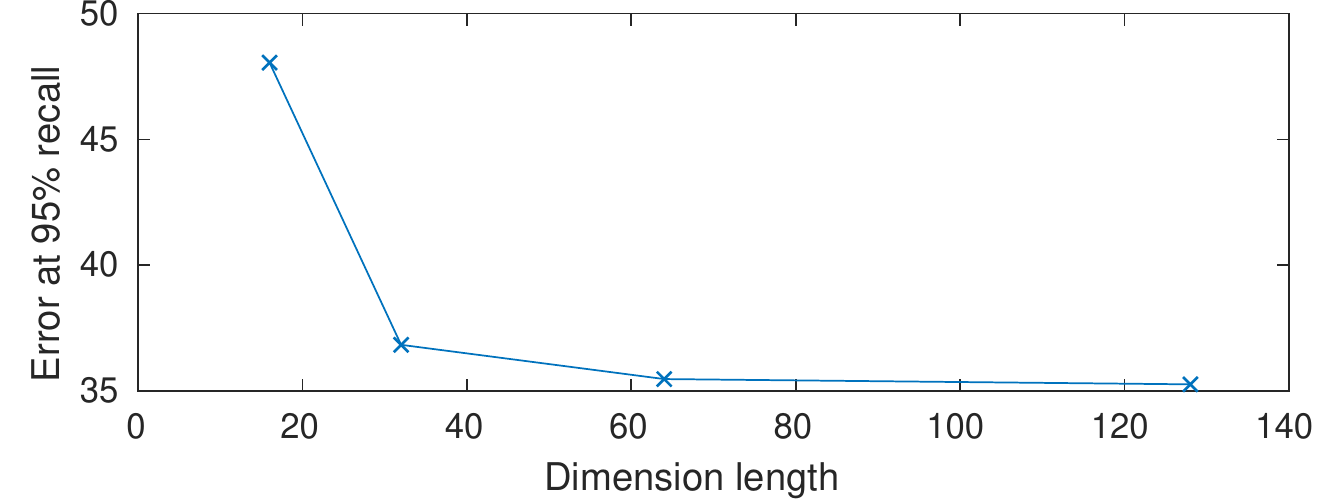}
\caption{Matching error at 95\% recall for different descriptor dimensionality}
\label{fig:dimension-ablation}
\end{figure}

 Fig. \ref{fig:dimension-ablation} shows the performance of our descriptor at different dimensionality. We observe that there is diminishing returns above 32 dimensions, and will use a feature dimension of $d=32$ for the rest of the experiments.

\begin{table}
\centering
\caption{Descriptor matching task. Error at 95\% recall. Lower is better.}
\label{table:descriptor-matching}
\begin{tabular*}{0.92\textwidth}{@{\extracolsep{\fill}}*{8}{c}}
\hline
 & SI\cite{SpinImage} & FPFH\cite{FPFH} & USC\cite{USC} & CGF\cite{CGF} & 3DMatch\cite{zeng20163dmatch} & PN++\cite{pointnetpp} & \textbf{Ours} \\
\hline
Error (\%) & 68.51 & 54.13 & 91.59 & 69.77 & 38.49 & 50.57 & \textbf{36.84} \\
\hline
\end{tabular*}
\end{table}

Table \ref{table:descriptor-matching} compares our descriptor with the baseline algorithms. Our learned descriptor yields a lower error rate than all other descriptors despite having a similar or smaller dimension. It performs significantly better than the best handcrafted descriptor (FPFH) which uses explicitly computed normals. The other two handcrafted descriptors (USC and SI), as well as the learned CGF consider the number of points falling in each subregion around the keypoint and could not differentiate the sparse point cloud clusters.
3DMatch performed well in differentiating randomly sampled clusters, but requiring a larger feature dimension.
Lastly, the modified PointNet++ network did not learn a good descriptor in the weakly supervised setting and gives significantly higher error than our descriptor despite having the same descriptor network structure.

\subsection{Keypoint Detection and Feature Description}

\begin{figure*}[t]
\centering
\includegraphics[height=4.2cm]{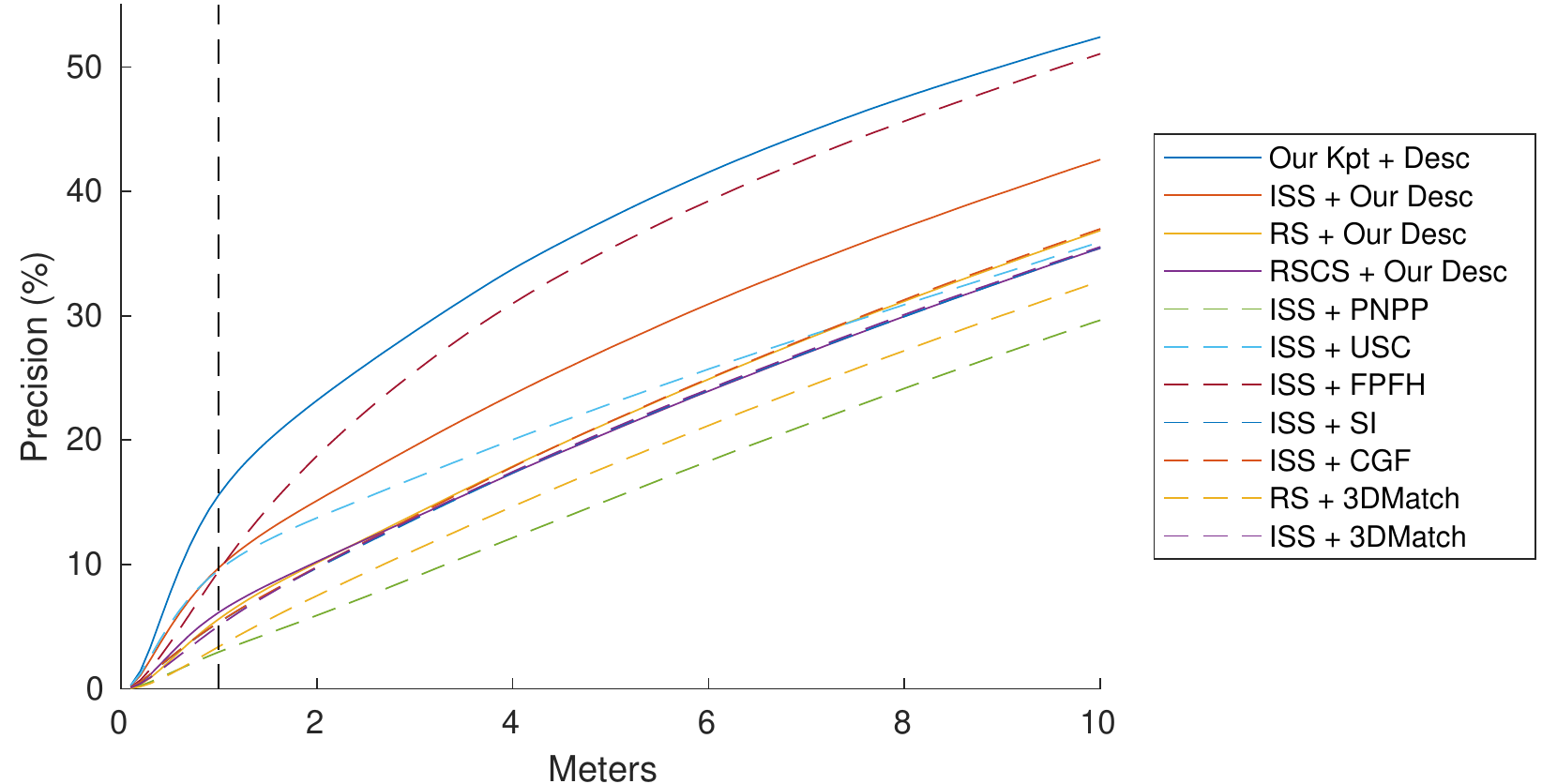}
\caption{Precision on the Oxford Robotcar dataset using different keypoint and descriptor combination. (Kpt = keypoints, Desc = descriptors)}
\label{fig:oxford-matchPrec}
\end{figure*}

\begin{figure*}[t]
\centering
\includegraphics[width=\textwidth]{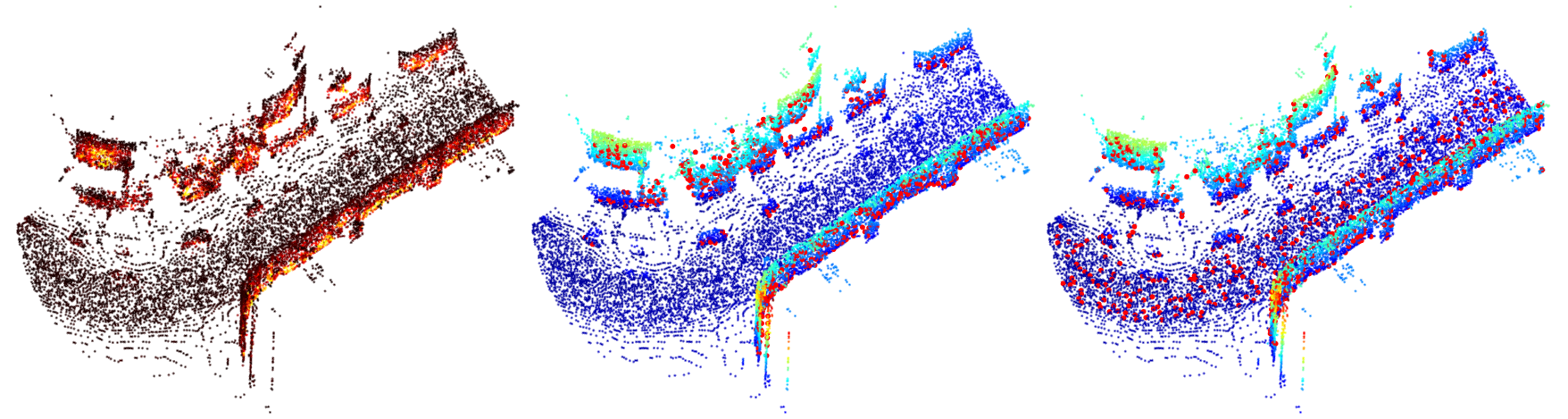}
\caption{Left: Attention using proposed method (brighter colors indicate higher attention). Middle: Our Keypoints (red dots). Right: ISS keypoints (red dots). Colors in middle and right images indicate different heights above ground.}
\label{fig:attention}
\end{figure*}

We follow the procedure in \cite{CGF} to evaluate the joint performance of keypoints and descriptors. For each keypoint descriptor in the first model of the pair, we find its nearest neighbor in the second model via exhaustive search. We then compute the distance between this nearest neighbor and its ground truth location.
We obtain a precision plot, as shown in Fig. \ref{fig:oxford-matchPrec}, by varying the distance threshold $x$ for considering a match as correct and plotting the proportion of correct matches.
We evaluate on all the baseline descriptors, and also show the performance of our descriptor with the ISS detector, random sampling of points (RS), and points obtained using Random Sphere Cover Set (RSCS) \cite{Elbaz2017CVPR}.
We tuned the cluster sizes for all baseline descriptors, as many of them require larger cluster sizes to work well. Nevertheless, our keypoint detector and descriptor combination still yields the best precision for all distance thresholds, and obtains a precision of 15.6\% at 1m. We also note that our descriptor underperforms when used with the two random sampling methods or the generic ISS detector, indicating the importance of a dedicated feature detector.

\medskip
\noindent{\textbf{Analysis of Keypoint Detection}} \enskip Fig. \ref{fig:attention} shows the attention weights computed by our network, as well as the retrieved keypoints. We also show the keypoints obtained using ISS for comparison. Our network learns to give higher attentions to lower regions of the walls (near the ground), and mostly ignores the ground and the cars (which are transient and not useful for matching). In comparison, ISS detects many non-distinctive keypoints on the ground and cars.

\subsection {Geometric Registration}
We test our keypoint detection and description algorithm on the geometric registration problem. We perform nearest neighbor matching on the computed keypoints and descriptors, and use RANSAC on these nearest neighbor matches to estimate a rigid transformation between the two point clouds. The number of RANSAC iterations is automatically adjusted based on 99\% confidence but is limited to 10,000. No subsequent refinement, e.g. using ICP is performed.
We evaluate the estimated pose against its ground truth using the Relative Translational Error (RTE) and Relative Rotation Error (RRE) as in  \cite{Elbaz2017CVPR,RTERRE}. 
We consider registration successful when the RTE and RRE are both below a predefined threshold of 2m and 5$^\circ$, and report the average RTE and RRE values for successful cases. We also report the average number of RANSAC iterations.

\begin{table}
\centering
\caption{Performance on the Oxford RobotCar dataset.}
\label{table:model-reg}
\begin{tabular*}{\textwidth}{@{\extracolsep{\fill}}c||cccc}
\hline
Method & RTE (m) & RRE ($^\circ$) & Success Rate & Avg \# iter\\
\hline
ISS \cite{ISS} + FPFH \cite{FPFH} & $0.396\pm 0.290$ & $1.60\pm 1.02$ & 92.32\% & 7171 \\
ISS \cite{ISS} + SI \cite{SpinImage} & $0.415\pm 0.309$ & $1.61\pm 1.12$ & 87.45 \% & 9888\\
ISS \cite{ISS} + USC \cite{USC} & $0.324\pm 0.270$ & $1.22\pm 0.95$ & 94.02\% & 7084 \\
ISS \cite{ISS} + CGF \cite{CGF}  & $0.431\pm 0.320$ & $1.62\pm 1.10$ & 87.36\% & 9628 \\
RS + 3DMatch \cite{zeng20163dmatch}  & $0.616\pm 0.407$ & $2.02\pm 1.17$ & 54.64\% & 9848 \\
ISS \cite{ISS} + 3DMatch \cite{zeng20163dmatch}  & $0.494\pm 0.366$ & $1.78\pm 1.21$ & 69.06\% & 9131 \\
ISS \cite{ISS} + PN++ \cite{pointnetpp} & $0.511\pm 0.391$ & $1.88\pm 1.20$ & 48.86\% & 9904 \\
\hline
RS + Our Desc & $0.435\pm 0.305$ & $1.64\pm 1.04$ & 90.28\% & 9941 \\
RSCS \cite{Elbaz2017CVPR} + Our Desc & $0.386\pm 0.292$ & $1.46\pm 1.01$ & 92.64\% & 9913\\
ISS \cite{ISS} + Our Desc & $0.314\pm 0.262$ & $1.08\pm 0.83$ & 97.66\% & 7127 \\
\textbf{Our Kpt + Desc} & $\mathbf{0.300\pm 0.255}$ & $\mathbf{1.07\pm 0.85}$ & \textbf{98.10\%} & \textbf{2940} \\
\hline
\end{tabular*}
\end{table}

\subsubsection{Performance on Oxford RobotCar}
Table \ref{table:model-reg} shows the performance on the Oxford dataset.
We observe the following: (1) using a keypoint detector instead of random sampling improves geometric registration performance, even for 3DMatch which is designed for random keypoints, (2) our learned descriptor gives good accuracy even when used with handcrafted descriptors or random sampling, suggesting that it generalizes well to generic point clusters, (3) our detector and descriptor combination gives the highest success rates and lowest errors. This highlights the importance of designing a keypoint detector and descriptor simultaneously, and the applicability of our approach to geometric registration. 
Some qualitative registration results can be found in Fig. \ref{fig:matches-qualitative}(a).

\subsubsection{Performance on KITTI Dataset}
We evaluate the generalization performance of our network on the KITTI odometry dataset by comparing the geometric registration performance against ISS + FPFH in Table \ref{table:model-reg-kitti}. We use the same parameters as the Oxford dataset for all algorithms, and did not fine-tune our network in any way. Nevertheless, our 3DFeat-Net outperforms all other algorithms in most measures. It underperforms CGF slightly in terms of RTE, but has a significantly higher success rate and requires far fewer RANSAC iterations.
We show some matching results on the KITTI dataset in Fig. \ref{fig:matches-qualitative}(b).

\begin{table}[t]
\centering
\caption{Performance on the KITTI odometry dataset.}
\label{table:model-reg-kitti}
\begin{tabular*}{\textwidth}{@{\extracolsep{\fill}}c||cccc}
\hline
Method & RTE (m) & RRE ($^\circ$) & Success Rate & Avg \# iter\\
\hline
ISS \cite{ISS} + FPFH \cite{FPFH} & 
    $0.325\pm 0.270$           &  $1.08\pm 0.82$           & 58.95\%          &  7462 \\
ISS \cite{ISS} + SI \cite{SpinImage} & 
    $0.358\pm 0.304$           & $1.17\pm 0.94$            & 55.92\%          &  9219\\
ISS \cite{ISS} + USC \cite{USC} & 
    $0.262\pm 0.275$           & $0.83\pm 0.75$            & 78.24\%          &  7873 \\
ISS \cite{ISS} + CGF \cite{CGF} &
    $\mathbf{0.233\pm 0.266}$           &  $0.69\pm 0.60$           & 87.81\%          &  7442 \\
RS + 3DMatch \cite{zeng20163dmatch} &
    $0.377\pm 0.298$           &  $1.21\pm 0.84$           & 83.96\%          &  8674 \\
ISS \cite{ISS} + 3DMatch \cite{zeng20163dmatch} & 
    $0.283\pm 0.272$           &  $0.79\pm 0.65$           & 89.12\%          &  7292 \\
\hline
\textbf{Our Kpt + Desc} & 
    $0.259\pm 0.262$  &  $\mathbf{0.57\pm 0.46}$  & \textbf{95.97\%} &  \textbf{3798} \\
\hline
\end{tabular*}
\end{table}

\subsubsection{Performance on ETH Dataset}
\begin{figure*}[t]
    \centering
    \includegraphics[width=0.95\textwidth]{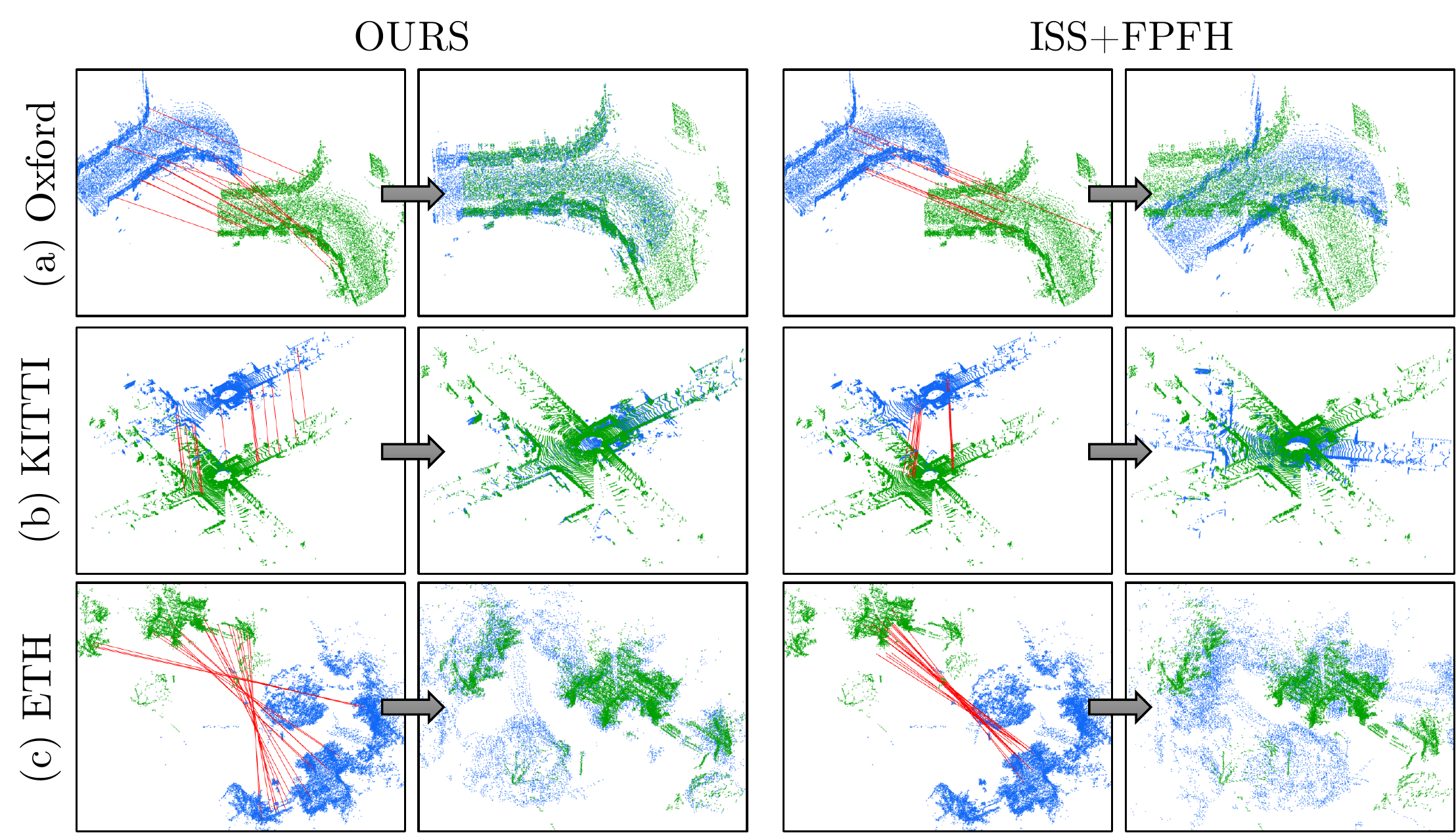}
    \caption{Qualitative registration results, using our approach (left) and ISS + FPFH (right). We only show a random subset of matches retained after RANSAC, and exclude the ground in (c) for clarity. We also show the results using ISS + FPFH.}
    \label{fig:matches-qualitative}
\end{figure*}

We compare our performance against LORAX \cite{Elbaz2017CVPR}, which evaluates on 9 models from the Gazebo dataset and 3 models from the Wood dataset. We show our performance on the best performing point clouds for each algorithm in Table \ref{table:model-reg-eth} since it is not explicitly stated in \cite{Elbaz2017CVPR} which datasets were used. Note that success rate in this experiment refers to a RTE of below 1m to be consistent to \cite{Elbaz2017CVPR}. We also report the success rate over the entire dataset. Detailed results on the entire dataset can be found in the supplementary material.
LORAX considers 3 best descriptor matches for each keypoint. These matches are used to compute multiple pose hypotheses which are then refined using ICP for robustness. For our algorithm and baselines, we only consider the best match, compute a single pose hypothesis and do not perform any refinement of the pose. Despite this, our approach outperforms \cite{Elbaz2017CVPR} and most baseline algorithms. It only underperforms USC which uses a much larger dimension (1980D).
Fig. \ref{fig:matches-qualitative}(c) shows an example of a successful matching by our approach.

\begin{table}
\centering
\caption{Performance on the ETH dataset, the results of FPFH and LORAX are taken from \cite{Elbaz2017CVPR}. The last column indicates the success rate over the entire dataset.}
\label{table:model-reg-eth}
\begin{tabularx}{\textwidth}{@{\extracolsep{\fill}}c||ZZZ|>{\hsize=.6\hsize}Z}
\hline
Method & RTE (m) & RRE ($^\circ$) & Success Rate & (All) \\
\hline
FPFH \cite{FPFH} & 
    $0.44\pm 0.2$              & $12.2\pm 4.8$             &  67\%             &  - \\
LORAX \cite{Elbaz2017CVPR} & 
    $0.42\pm 0.27$             & $2.5\pm 1.2$              &  83\%             &  - \\
\hline
ISS \cite{ISS} + SI \cite{SpinImage} & 
    $0.176\pm 0.083$           &  $1.97\pm 0.74$           &  \textbf{100\%}   &  93.7\%  \\
ISS \cite{ISS} + USC \cite{USC} & 
$\mathbf{0.130\pm 0.056}$  &  $1.52\pm 0.30$  & \textbf{100\%}    & \textbf{100}\%  \\
ISS \cite{ISS} + CGF \cite{CGF} & 
    $0.157\pm 0.066$           &  $\mathbf{1.47\pm 0.60}$  &  \textbf{100\%}   &  92.1\%  \\
RS + 3DMatch \cite{zeng20163dmatch} & 
    $0.292\pm 0.199$           &  $4.71\pm 3.16$           &  91.7\%           &  33.3\% \\
ISS \cite{ISS} + 3DMatch \cite{zeng20163dmatch} & 
    $0.401\pm 0.222$           &  $5.32\pm 3.25$           &  \textbf{100\%}   &  33.3\% \\
\hline
\textbf{Our Kpt + Desc} & 
    $0.156\pm 0.112$           &  $1.56\pm 0.66$           &  \textbf{100\%}   &  95.2\% \\     
\hline
\end{tabularx}
\end{table}
\FloatBarrier
\section{Conclusion}
We proposed the 3DFeat-Net model that learns the detection and description of keypoints in Lidar point clouds in a weakly supervised fashion by making use of a triplet loss that takes into account individual descriptor similarities and the saliency of input 3D points. Experimental results showed our learned detector and descriptor compares favorably against previous handcrafted and learned ones on several outdoor gravity aligned datasets.
However, we note that our network is unable to train well on overly noisy point clouds, and the use of PointNet limits the max size of the input point cloud. 
We also do not extract a fully rotational invariant descriptor. We leave these as future work.

\clearpage

\bibliographystyle{splncs04}

\end{document}